\pdfoutput=1

\documentclass[11pt]{article}

\usepackage{EMNLP2022}

\usepackage{times}
\usepackage{latexsym}
\usepackage{makecell}
\usepackage{xurl}

\usepackage[T1]{fontenc}

\usepackage[utf8]{inputenc}

\usepackage{microtype}

\usepackage{enumitem}

\usepackage{inconsolata}

\usepackage{amsmath,amsfonts,amssymb,bm}
\usepackage{pifont} 
\usepackage{graphicx,subfig,epsfig,fancybox} 
\usepackage{float}
\usepackage{color} 
\usepackage[ruled,vlined]{algorithm2e} %
\usepackage{algorithmic}
\usepackage{multirow}
\usepackage{comment}
\usepackage{hyperref}
\usepackage{natbib} 

\renewcommand{\vec}[1]{\boldsymbol{#1}}

\usepackage{adjustbox}
\usepackage{array}
\usepackage{booktabs}

\newcolumntype{R}[2]{%
    >{\adjustbox{angle=#1,lap=\width-(#2)}\bgroup}%
    l%
    <{\egroup}%
}

\usepackage{trackchanges}
\addeditor{JZ}
\addeditor{YL}
\addeditor{LW}

\newcommand{\dataset}{ActDial}
\newcommand{\method}{FlowEval}

\title{FlowEval: A Consensus-Based Dialogue Evaluation Framework Using Segment Act Flows}

\author{Jianqiao Zhao$^1$\thanks{\ \ Equal contributions. Wanyu participated in building the dataset, while doing her internship with Prof. Liwei Wang.}\ , Yanyang Li$^1$\footnotemark[1]\ , Wanyu Du$^3$\footnotemark[1]\ , Yangfeng Ji$^3$, \\
\textbf{Dong Yu$^4$, Michael R. Lyu$^1$, Liwei Wang$^{12}$}\thanks{\ \ Corresponding author.} \\
$^1$Department of Computer Science and Engineering, The Chinese University of Hong Kong \\
$^2$Shanghai AI Laboratory
$^3$Department of Computer Science, University of Virginia \\
$^4$Tencent AI Lab, Bellevue \\
\{jqzhao, yyli21, lyu, lwwang\}@cse.cuhk.edu.hk 
}

\begin{document}
\maketitle

\begin{abstract}
Despite recent progress in open-domain dialogue evaluation, how to develop automatic metrics remains an open problem.
We explore the potential of dialogue evaluation featuring dialog act information, which was hardly explicitly modeled in previous methods.
However, defined at the utterance level in general, dialog act is of coarse granularity, as an utterance can contain multiple segments possessing different functions.
Hence, we propose \emph{segment act}, an extension of dialog act from utterance level to segment level, and crowdsource a large-scale dataset for it.
To utilize \emph{segment act flows}, sequences of segment acts, for evaluation, we develop the first consensus-based dialogue evaluation framework, \emph{\method{}}.
This framework provides a reference-free approach for dialog evaluation by finding pseudo-references. 
Extensive experiments against strong baselines on three benchmark datasets demonstrate the effectiveness and other desirable characteristics of our \emph{\method{}}, pointing out a potential path for better dialogue evaluation.




\end{abstract}

\section{Introduction}
\label{sec:introduction}

Dialogue evaluation plays a crucial role in the recent advancement of dialogue research.
While human evaluation is often considered as a universal and reliable method by the community \citep{Smith2022HumanEO}, automatic dialogue evaluation metrics draw growing attention as they can assess dialogues with faster speed and lower cost \citep{DBLP:conf/aaai/TaoMZY18, DBLP:conf/emnlp/HuangYQLL20, DBLP:journals/corr/fed}.

Traditional word-overlap metrics, like BLEU \citep{papineni-etal-2002-bleu} and METEOR \citep{banerjee-lavie-2005-meteor}, lose some of their effectiveness in the dialogue setting as reliable references are hard to obtain \citep{liu-etal-2016-evaluate}.
Recent works tackle this problem by leveraging more sophisticated architectures \citep{DBLP:journals/corr/dynaeval, li-etal-2021-conversations} and harnessing the power of large models \citep{DBLP:journals/corr/fed}.
Although these recent metrics claim to show some progress towards higher correlation with humans, the gap between automatic metrics and human evaluation is still noticeable \citep{Yeh2021comprehensive}.
Automatic open-domain dialogue evaluation is still an open question, and extensive efforts have been made to improve performance from different angles \citep{pang-etal-2020-towards-holistic, DBLP:journals/corr/predictive-engagement, mehri-eskenazi-2020-usr, phy-etal-2020-deconstruct}.

\begin{table}[t!]
  \centering
  \setlength{\tabcolsep}{5pt}
  \small
  \begin{tabular}{p{4.5cm}|p{2.5cm}}
    \toprule
    \makecell[c]{\textbf{Dialogue Section}} & \makecell[c]{\textbf{Segment Act Flow}} \\
    \midrule
    \textbf{Speaker1}: \colorbox{orange}{How are you?} \colorbox{green}{May I have a cup of coffee?} & \colorbox{orange}{\texttt{greeting}}, \colorbox{green}{\texttt{directive}} \\
    \midrule
    \textbf{Speaker2}: \colorbox{yellow}{Hmm.} \colorbox{cyan}{Certainly.} \colorbox{pink}{What kind of coffee do you like?} \colorbox{lime}{We have espresso and latte.} & \colorbox{yellow}{\texttt{backchannel-}} \colorbox{yellow}{\texttt{success}}, \colorbox{cyan}{\texttt{commissive}}, \colorbox{pink}{\texttt{question}}, \colorbox{lime}{\texttt{inform}} \\
    \bottomrule
  \end{tabular}
  \caption{\label{table:dialog-example}
  A snippet of an open-domain dialogue and its segment act flow. Each segment is marked with the same color as its corresponding segment act label.
  }
  \vspace{-0.2cm}
\end{table}

Among those newly released metrics \citep{DBLP:journals/corr/dynaeval, DBLP:journals/corr/fed, li-etal-2021-conversations, pang-etal-2020-towards-holistic, DBLP:conf/aaai/TaoMZY18, mehri-eskenazi-2020-usr, phy-etal-2020-deconstruct}, hardly any explicitly employs dialog act, one of the pillars of dialogue study, in their methods.
Intuitively, introducing dialog act into open-domain dialogue evaluation should be beneficial: 
a sequence of dialog acts distills the core function of each utterance and can potentially reveal how speakers interact in general.
However, directly using preexisting dialog act definitions \cite{DBLP:journals/corr/switchboard, hemphill-etal-1990-atis} seems undesirable, as an utterance can contain several segments that possess different conversational functions.
We show our observation in Table~\ref{table:dialog-example}.
By saying ``\textit{Hmm. Certainly. What kind of coffee do you like? We have espresso and latte.}'', the participant first acknowledges the conversation with backchannel, then commits to finish the request by saying ``\textit{Certainly}''.
Later, the speaker asks for a more concrete order and offers all the options.
In human conversations, it is common to have more than one function for each utterance, which means using a single dialog act to express the core function of an utterance inevitably suffers from information loss.
To solve this issue, we extend the concept of dialog act from the utterance level to the segment level.
We name these segment level dialog act \emph{segment act} and a sequence of segment acts a \emph{segment act flow}.

One difficulty of using segment act for open-domain dialogue evaluation is the lack of related data.
Since there is no dataset for segment act, we follow the ISO 24617-2 annotation criteria \citep{bunt2019dialogbank} and propose a simplified ISO-format segment act tagset.
We crowdsource large-scale segment act annotations on two popular open-domain dialogue datasets: ConvAI2 \citep{dinan2020second} and DailyDialog \citep{li2017dailydialog}. 
We name our dataset \emph{\dataset{}}.

Another challenge of incorporating segment act into open-domain dialogue evaluation lies in finding a suitable way to assess dialogues with the segment act feature.
Modeling segment act flow is not trivial. 
On the one hand, dialogues have different numbers of turns and, thus, have varying lengths of segment act sequences.
On the other hand, defining and finding the ground-truth segment act flow for a dialogue are almost infeasible, discouraging the development of any reference-based methods.
To overcome this challenge, we design the first consensus-based reference-free open-domain dialogue evaluation framework, \emph{\method{}}. 

For a dialogue to be evaluated, our \emph{\method{}} first obtains the segment act flow, e.g., from a trained classifier.
Then, we harvest segment act features, from a dedicated BERT-like \citep{devlin-etal-2019-bert} masked segment act model, and content features, from RoBERTa-large \citep{DBLP:journals/corr/abs-1907-11692}.
We retrieve pseudo-references from the training set, according to the segment act features as well as content features. 
Last, we evaluate the dialogue with the consensus of the pseudo-references, fusing metrics from both segment act and word-overlap perspectives.
The essence of our consensus-based framework lies in retrieving pseudo-references and using the consensus of pseudo-references to assess a new dialogue.
This process can be regarded as reference-free, since no additional dialogue evaluation label is required.
Not limited to segment act feature, our proposed consensus-based framework is compatible to a wide range of features and metrics, such as sentiment features, engagingness feature, etc.


Extensive experiments are carried out against the state-of-the-art baselines 
on Controllable Dialogue dataset \citep{see2019makes}, FED dataset \citep{DBLP:journals/corr/fed}, and DSTC9 dataset \citep{DBLP:journals/corr/dstc9}.
The result supports that segment act flow is effective in dialogue evaluation:
our consensus-based method achieve the best or comparable correlation with human evaluation. Additionally, segment act flow can bring complementary information to metrics that heavily focus on the raw text of dialogues.

In summary, the contributions of this work are three-fold:
\begin{enumerate}[noitemsep,nolistsep]
\item We propose to model the segment level act as the dialog flow information for open-domain dialogue evaluation. 

\item We are the first to propose a consensus-based framework for open-domain dialogue evaluation. 
Our studies show that the consensus approach can work efficiently even when the size of the search set, i.e., the number of dialogues in the training set, is around ten thousand. This attainable size shows the promise of our consensus approach for dialogue evaluation and other natural language evaluation tasks.

\item Our method can reach the best or comparable performance when compared with state-of-the-art baselines.
Additional experiments are conducted to examine detailed properties of our method and consensus process.
\end{enumerate}

We will release all code and data once the paper is made public.


\section{Related Works}
\label{sec:related-works}
\subsection{Automatic Dialog Evaluation Metrics}
\label{related:auto-metrics}

RUBER \citep{DBLP:conf/aaai/TaoMZY18} combines a reference-based metric and a reference-free metric where the reference-free metric is learned by an RNN-based model to judge if a response is appropriate for the dialogue history.
GRADE \citep{DBLP:conf/emnlp/HuangYQLL20} adopts the graph structure to represent dialogue topics and enhances the utterance-level contextualized representations with topic-level graph representations to better evaluate the coherence of dialogues.
DynaEval \citep{DBLP:journals/corr/dynaeval} reaches the highest correlation with human evaluation on FED dataset \citep{DBLP:journals/corr/fed}, by utilizing the Graph Convolutional Network to capture the dependency between dialogue utterances.
FED \citep{DBLP:journals/corr/fed} measures 18 different qualities of dialogues by computing the likelihood of DialoGPT \citep{DBLP:journals/corr/dialogpt} generating corresponding handwritten follow-up utterances.
In addition to commonly-used dialogue context, Flow score \citep{li-etal-2021-conversations} takes into account the semantic influence brought by each utterance, which is defined to be the difference between the dense representations of two adjacent dialogue histories. 
Flow score employs special tasks to promote a better modeling of semantic influence during pretraining.
It correlates best with human evaluation on the DSTC9 dataset \citep{DBLP:journals/corr/dstc9}.

Different from Flow score and other related works, our method explicitly models the segment acts of a dialog, which deliver clear and interpretable functions for each utterance segment, rather than dense representations.


\subsection{Dialog Act in Dialogue Systems}
\label{related:segment-act}
Dialog act \citep{DBLP:journals/corr/switchboard, shriberg-etal-2004-icsi} and similar concepts, like intent \citep{hemphill-etal-1990-atis, larson-etal-2019-intent-oos}, have been widely studied in the past decades.
\citet{walker-passonneau-2001-date} construct a dialog act tagging scheme and evaluate travel planning systems \cite{Walker2001DARPACD} based on standalone dialog acts, rather than dialog act sequences.
This tagging scheme, while provides more detailed information compared with previous works, only focuses on the system side in a task-oriented setting and may need major modifications when applied to open-domain dialogues.
After the initial flourish, recent works come with their own purposes and tagsets for dialog act, tailored for different scenarios or special needs \citep{DBLP:journals/corr/multiwoz, DBLP:journals/corr/midas,DBLP:journals/corr/DA-coherence}.

In this work, we propose segment act, an extension of dialog act to the utterance segment level, and design its corresponding tagset. Our segment-focused arrangement can not only cover the diverse scenarios of open-domain dialogues, but also provide finer-grained information for dialogue evaluation than prevailing dialog act designs.

\subsection{Consensus-Based Methods}
\label{related:consensus}
Consensus-based methods have been adopted in image captioning \citep{DBLP:journals/corr/DevlinGGMZ15,wang2017diverse,deshpande2019fast} and evaluation \citep{DBLP:journals/corr/cider, DBLP:journals/corr/CIDEr-R}.

\citet{DBLP:journals/corr/DevlinGGMZ15} retrieve nearest neighbors in the sense of image feature and use them as consensus.
They later take the caption that has the best word overlap with the consensus as the generation results and achieve competitive performance against other caption generation techniques.
Consensus-based Image Description Evaluation (CIDEr) \cite{DBLP:journals/corr/cider} measures the similarity of a generated caption against a set of human-written sentences using a consensus-based protocol. 
Our proposed method shares similar element as it also involves evaluating by the closeness to a consensus of human sentences.
However, to the best of our knowledge, this is the first work that adapts the consensus-based evaluation to dialogues.

\section{\dataset{}: A Segment Act Dataset on Open-Domain Dialogues}
\label{sec:dataset}

We propose the new concept of segment act, extracting the core function of each segment in an utterance.
We then crowdsource a large-scale open-domain dialogue dataset with our proposed segment act labels, called \emph{\dataset{}}.


\subsection{Our Segment Act Tagset}
We design an open-domain segment act tagset based on the ISO 24617-2 annotation criteria \citep{bunt2019dialogbank}.
We define a segment act as a functional label that expresses the communicative goal of participants in a conversation, which is irrelevant to syntactic or sentiment details. 
Based on this definition, we conduct combination operations, like merging \texttt{Choice-Question}, \texttt{Check Question}, etc. into \texttt{question}, on the original 56 labels proposed by \citet{bunt2019dialogbank} and eventually obtain 11 labels as our tagset.
These combination operations guarantee a robust coverage on diverse dialogue expressions and mutual exclusiveness between different segment act labels.
From our later experiments, these 11 labels capture key information from dialogues while remain simple enough to enable large-scale accurate annotations. Detailed definition and examples of each segment act can be found in Appendix \ref{sec:appendix-intent-definition}

\subsection{Datasets and Segmentation}
\label{subsec:dial-seg}
We crowdsourced segment act annotations on ConvAI2 \citep{dinan2020second} and DailyDialog \citep{li2017dailydialog}.
The details of the crowdsourcing process are in Appendix \ref{sec:appendix-dataset-crowdsource}

The \textbf{ConvAI2} dataset
is based on the PersonaChat dataset \citep{zhang-etal-2018-personalizing}, where all dialogues are constructed by asking two crowd-workers to chat with each other based on randomly assigned persona profiles.
ConvAI2 is a widely-used benchmark for many state-of-the-art dialogue systems \citep{golovanov2019large,bao2019plato,shuster2019dialogue,roller2020recipes}. 

The \textbf{DailyDialog} dataset \cite{li2017dailydialog}
is constructed by crawling raw data from various English-learning websites.
Note that DailyDialog already has 4 dialog act labels: \texttt{question}, \texttt{inform}, \texttt{directive}, and \texttt{commissive}.
Our finer-grained annotation which takes social chit-chat and simple feedback into account can better cover diverse dialogue scenarios and provide  extra information.

Following our definition of segment acts, we split each utterance into multiple segments using NLTK \citep{bird-loper-2004-nltk} sentence punctuation tokenizer \citep{kiss-strunk-2006-unsupervised}.
The resulting segments will have their own segment act labels during annotation.
Each segment is annotated by three different crowd-workers.
With our special tagset design and the segmentation process, annotators can easily reach substantial agreement and deliver a high-quality dataset: 
Fleiss' kappa \citep{fleiss1971measuring} achieves 0.754 for DailyDialog and 0.790 for ConvAI2.
Detailed statistics of our \dataset{} dataset is documented in Appendix \ref{sec:appendix-dataset-statistics}.
Note that the majority of the segments are labeled as \texttt{question} and \texttt{inform}.
This is common in dialog act datasets \citep{DBLP:journals/corr/switchboard,DBLP:journals/corr/midas} as most of the dialogues consist of asking for information and stating fact or opinion.

\section{\method{}: A Segment-Act-Flow Aware Evaluation Metric}
\label{sec:methods}
In this section, we describe the details of our proposed dialogue evaluation framework, \emph{\method{}}.
\method{} is implemented in three stages: segment act harvesting, retrieval, and assessment.

\subsection{Segment Act Harvesting}
\label{sub-sec:annotation}
In order to utilize the segment act flow, we first need to harvest the segment act labels for an unseen raw dialogue $U$.
In our experiments unless specified, the segment act labels are acquired by a text classification model, based on RoBERTa-large \cite{DBLP:journals/corr/abs-1907-11692} and fine-tuned on \dataset{}.
The accuracy of this classifier is 90\% on unseen data.
In the end, we will have the annotated segment act flow $A_U=\{a_1,\cdots,a_i,\cdots,a_n\}$ for the dialogue $U$, where $a_i$ is the segment act label for $i$-th segment and $n$ is the number of segments in $U$.

\begin{figure*}
\begin{center}
    \includegraphics[scale=0.84]{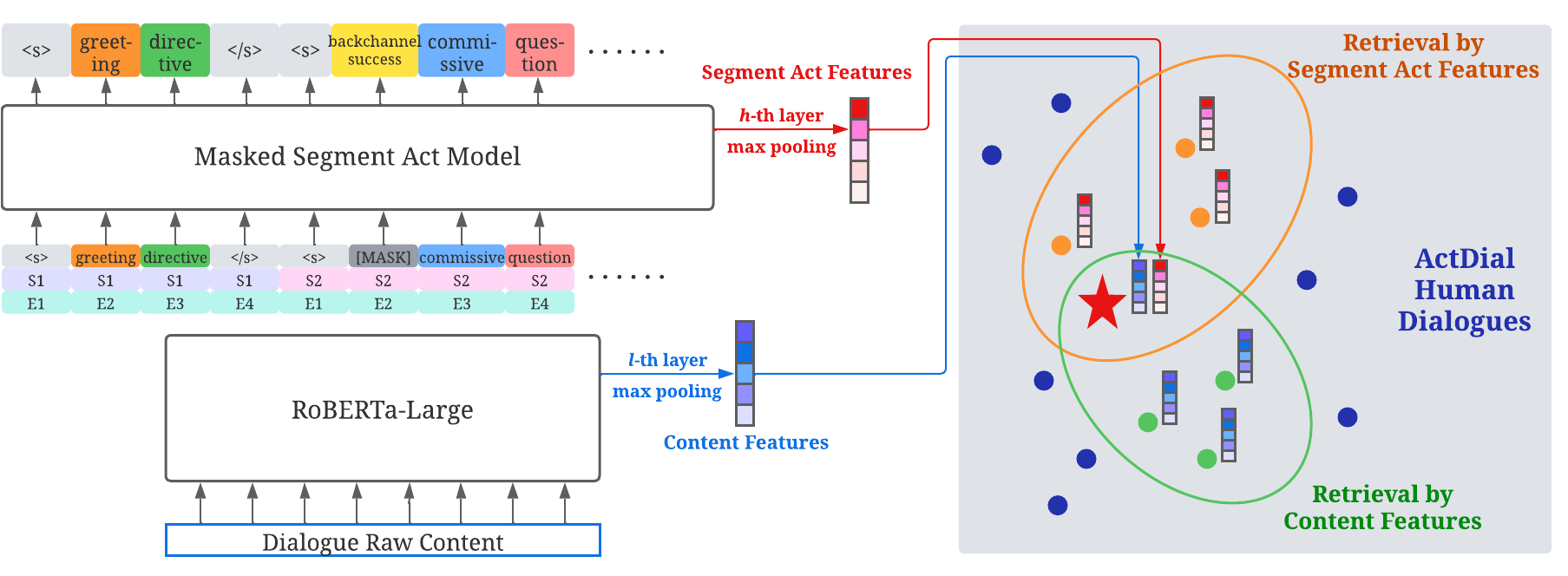}
\end{center}
\caption{
\label{fig:retrieval}
Extract segment act and content features. Retrieve closest human dialogues from \dataset{} dataset.
}
\end{figure*}

\subsection{Retrieval}
\label{sec:retrieval}

For the retrieval process, \method{} retrieves two sets of dialogues based on segment act features and content features respectively.
The search space for \method{} is our \dataset{} dataset and the unseen raw dialogue $U$ serves as query. 
\method{} first extracts segment act features from a masked segment act model, and retrieves $k^a$ nearest neighbors for $U$ based on our defined similarity function.
Then, \method{} extracts content features from a RoBERTa-large model, and retrieves $k^c$ nearest neighbours for $U$ based on another similarity function.
The final outcome of this retrieval stage is $k=k^a+k^c$ relevant dialogues for the unseen dialogue $U$.
Figure~\ref{fig:retrieval} illustrates this process in detail.

\paragraph{Segment Act Flow Features.}
To extract segment act flow features, we treat every segment act label as a word and a segment act flow of a dialogue as a sequence.
We then train a masked language model \cite{devlin-etal-2019-bert} called \emph{ActBERT} on all segment act flows in our \dataset{} datasets.
Detailed implementation of ActBERT is documented at Appendix \ref{sec:appendix-actbert}.
ActBERT has an accuracy of 81\% for predicting the masked segment act on unseen data, which is significantly higher than guessing the majority segment act label (67\%).
This means that our ActBERT indeed captures reliable features from the segment act flow.
ActBERT will be used to extract segment act features for any dialogue that has segment act flow.

Given a dialogue $D$, we first pass $D$'s segment act flow $A_D$ into ActBERT.
The output of $h$-th intermediate layer of ActBERT, $H^h_D\in\mathbb{R}^{n\times d}$, will be chosen, where $h$ is a hyper-parameter, $n$ is the number of segments in $D$ and $d$ is the hidden size of ActBERT.
$H^h_D$ is then max-pooled along the $n$ dimension to construct a fixed length vector $\bar{H}^h_D\in\mathbb{R}^{d}$ as the segment act feature of $D$.

We further employ TF-IDF features to constrain the retrieved dialogues to have a similar topic as $U$.
We collect the word count statistics from our \dataset{} dataset and compute the TF-IDF feature vector $T_D\in\mathbb{R}^{v}$ for any dialogue $D$, where $v$ is the vocabulary size.

Having the feature set $\{\bar{H}^h_U,T_U\}$ of $U$ and $\{\bar{H}^h_R,T_R\}$ of a human dialogue $R$ in \dataset{}, we define an segment-act-based similarity metric $S^a$ to retrieve $k^a$ nearest neighbors $\{R_i\}_{k^a}$:
\begin{equation}
\label{eqn:act}
S^a(U,R)=(1+\mathrm{cos}(\bar{H}^h_U,\bar{H}^h_R))(1+\mathrm{cos}(T_U,T_R))
\end{equation}
where $\mathrm{cos}$ is the cosine similarity.
$S^a$ in Eq.~\ref{eqn:act} will only score high if $R$ has a segment act flow as well as a topic closed to $U$.

\paragraph{Content Features.}
Retrieval with segment act features only might miss dialogues that discussed similar contents as $U$ but speakers communicated in a different way to $U$.
Therefore, we retrieve from \dataset{} again but using features with regard to the content of $U$.

We use RoBERTa-large \cite{DBLP:journals/corr/abs-1907-11692}, a pre-trained language models, to extract the content feature of any dialogue $D$.
We first feed the raw text of $D$ into RoBERTa and take the $l$-th layer representation $L^l_D\in\mathbb{R}^{m\times d}$ of RoBERTa, where $l$ is a hyper-parameter, $m$ is the number of tokens in $D$ and $d$ is the hidden size of RoBERTa.
$L^l_D$ is then max-pooled along the $m$ dimension to obtain a fixed-length content feature vector $\bar{L}^l_D\in\mathbb{R}^{d}$ for $D$.
Having the content feature $L^l_U$ of $U$ and $L^l_R$ of $R$ in \dataset{}, we define a content-based similarity metric $S^c$ for the second-round retrieval to retrieve $k_c$ nearest neighbors $\{R_i\}_{k^c}$:
\begin{equation}
\label{eqn:content}
S^c(U,R)=\cos (L^l_U,L^l_R)
\end{equation}
$S^c$ in Eq.~\ref{eqn:content} will output a high score if $R$'s content is closed to $U$.
The final retrieved set of dialogues will be $\{R_i\}_k=\{R_i\}_{k^a}\bigcup\{R_i\}_{k^c}$.

\subsection{Assessment}

We define a metric to find the closest $R^{*}\in\{R_i\}_k$ to $U$ by treating this small retrieved set $\{R_i\}_k$ as pseudo-references.
The distance between $R^{*}$ and $U$ will be the final score of $U$.
Concretely, we have the following scoring function $F$:
\begin{equation}
    \label{eqn:f}
    F(U)=\max_{R\in\{R_i\}_k}w F^a(U,R)+(1-w) F^c(U,R)
\end{equation}
\begin{align}
    \label{eqn:fi}
    F^a(U,R)&=S^a(U,R)\cdot\mathrm{BLEU}(A_U,A_R)\\
    \label{eqn:fc}
    F^c(U,R)&=\mathrm{BERTScore}(U,R)
\end{align}
where $w$ is a hyper-parameter between 0 and 1.
Eq.~\ref{eqn:f} assess $U$ from two aspects:
$F^a$, computed by Eq.~\ref{eqn:fi}, indicates whether speakers in $U$ interact naturally and is evaluated by ActBERT in Eq.~\ref{eqn:act} and BLEU score \cite{papineni-etal-2002-bleu} of the raw segment act flow $A_U$;
$F^c$, on the other hand, measures how natural sentences in $U$ are using BERTScore \cite{DBLP:conf/iclr/ZhangKWWA20} in Eq.~\ref{eqn:fc}.

\section{Experiments and Analysis}
\label{sec:experiments}

\begin{table*}[th]
  \centering
  \setlength{\tabcolsep}{4pt}
  \small
  \begin{tabular}{lccc|ccc|ccc}
    \toprule
    \makecell[c]{\multirow{2}{*}{\textbf{Metric}}} & \multicolumn{3}{c|}{\textbf{Controllable Dialogue}} & \multicolumn{3}{c|}{\textbf{FED Dataset}} & \multicolumn{3}{c}{\textbf{DSTC9 Dataset}} \\
    \cmidrule(lr){2-4}  \cmidrule(lr){5-7} \cmidrule(lr){8-10} 
    & Pearson & Spearman & Kendall & Pearson & Spearman & Kendall & Pearson & Spearman & Kendall \\
    \midrule
    BLEU & 0.132 & 0.136 & 0.104 & \multicolumn{6}{c}{\multirow{2}{*}{N/A}} \\
    \textsc{BERTScore} & \underline{0.282} & \underline{0.214} & \underline{0.162} \\
    \midrule
    \textsc{\makecell[l]{Consensus-\\BERTScore}} & 0.284 & 0.240 & 0.183 & 0.0874$^*$ & -0.037$^*$ & -0.023$^*$ & 0.060 & 0.054 & 0.039\\
    FED Metric & -0.025$^*$ & 0.010$^*$ & 0.007$^*$ & 0.134$^*$ & 0.126$^*$ & 0.088$^*$ & 0.117 & 0.108 & 0.078 \\
    DynaEval\_daily & 0.050$^*$ & 0.051$^*$ & 0.039$^*$ & 0.390 & 0.396 & 0.278 & 0.084 & 0.085 & 0.061 \\
    DynaEval\_emp &  0.026$^*$ &  -0.007$^*$ & -0.005$^*$ & \underline{0.464} & \underline{0.489} & \underline{0.341} & \textit{0.029}$^*$ & 0.046 & 0.033 \\
    Flow & -0.065$^*$ & -0.029$^*$ & -0.020$^*$ & -0.073$^*$ & -0.003$^*$ & -0.002$^*$ & \underline{0.154} & \underline{0.148} & \underline{0.106} \\
    \midrule
    FlowEval (Our) & 0.301 & \textbf{0.256} & \textbf{0.193} & 0.246 & 0.212 & 0.152 & 0.088 & 0.096 & 0.070 \\
    FlowEval (Our) + \underline{SOTA} & \textbf{0.327} & 0.250 & 0.190 & \textbf{0.468} & \textbf{0.493} & \textbf{0.342} & \textbf{0.165} & \textbf{0.161} & \textbf{0.116} \\
    FED Metric + \underline{SOTA} & 0.032$^*$ & 0.058$^*$ & 0.042$^*$ & 0.403 & 0.411 & 0.284 & 0.103 & 0.093 & 0.067\\
    DynaEval + \underline{SOTA} & 0.117$^*$ & 0.109$^*$ & 0.084$^*$ & \multicolumn{3}{c|}{N/A} & 0.054 & 0.059 & 0.042\\
    Flow + \underline{SOTA} & 0.207 & 0.140 & 0.107 & 0.460 & 0.471 & 0.327 & \multicolumn{3}{c}{N/A}\\
    \bottomrule
  \end{tabular}
  \caption{\label{tab:results}
  Correlations between different metrics and human evaluation on Controllable Dialogue (test set), FED and DSTC9 datasets. All values are statistically significant to $p$ < 0.05, unless that are marked by $^*$. \underline{SOTA} refers to the previous best performing methods (except our \method{}) in each dataset and is underlined.
  }
\end{table*}

\subsection{Benchmark Datasets}
\noindent\textbf{Controllable Dialogue Dataset} contains the human-to-bot conversation data collected by \citet{see2019makes}.
These conversations are based on the ConvAI2 dataset \citep{dinan2020second}.
We extend the original dataset by crowdsourcing segment act labels and human evaluation scores.
Details of human evaluation procedural are documented in Appendix \ref{sec:appendix-h2beval}.
There are 278 dialogues coming from 3 generative models.
28 dialogues are sampled randomly to form a validation set for hyperparameter tuning, while the rest make up the test set.

\noindent\textbf{FED Dataset} \citep{DBLP:journals/corr/fed} contains 125 human-to-bot conversations coming from three systems.
We take the mean of the 5 overall scores for each dialogue as the human evaluation score in our experiments.
We annotate all the segment act labels using the trained classifier described in Section~\ref{sub-sec:annotation}.

\noindent\textbf{DSTC9 Dataset} \citep{DBLP:journals/corr/dstc9} contains 2200 human-to-bot conversations from eleven chatbots.
We take the mean of the 3 human ratings as the final score.
All the segment act labels are predicted by a trained classifier.

\subsection{Methods}
\label{sub-sec:methods}
We describe all the baselines used for comparison and the implementation details of our method.\\
\noindent\textbf{FED metric} \citep{DBLP:journals/corr/fed}, leveraging the ability of DialoGPT-large \citep{DBLP:journals/corr/dialogpt} and the use of follow-up utterances, is an automatic and training-free evaluation method widely used by the community \citep{DBLP:journals/corr/dstc9}.\\ 
\noindent\textbf{DynaEval} \citep{DBLP:journals/corr/dynaeval} adopts the graph convolutionl network to model dialogues, where the graph nodes are dialogue utterances and graph edges represents the relationships between utterances.
DynaEval\_emp and DynaEval\_daily denote two variants trained on Empathetic Dialogues \citep{rashkin-etal-2019-towards} and DailyDialog \citep{li2017dailydialog} respectively.
DynaEval\_emp reaches the best correlation on FED dataset.\\
\noindent\textbf{Flow score} \cite{li-etal-2021-conversations}, considering the semantic influence of each utterance and modeling the dynamic information flow in dialogues, becomes the best evaluation method on DSTC9 dataset.\\
\noindent\textbf{BLEU} \citep{papineni-etal-2002-bleu} and \textbf{BERTScore} \citep{DBLP:conf/iclr/ZhangKWWA20} are two popular reference-based metrics.
The performance of BLEU and BERTScore are tested on Controllable Dialogue dataset only, as finding suitable reference is unfeasible on FED and DSTC9 dataset.
The process of how to find references on Controllable Dialogue and the implementation of BLEU and BERTScore are documented in Appendix \ref{sec:appendix-bleu&bertscore}.

\noindent\textbf{\method{} (our method)}
tune its hyperparameters on the validation set of Controllable Dialogue dataset and directly apply to the test set of Controllable Dialogue, FED and DSTC9.
Besides, since Controllable Dialogue dataset is constructed on top of ConvAI2 \cite{see2019makes}, we only use the DailyDialog part of \dataset{} for all the training and retrieval to prevent any data leakage.

\subsection{Results and Analysis}
The common practice to show the effectiveness of a dialogue evaluation metric is to calculate the Pearson, Spearman's, and Kendall correlation between human evaluation and the automatic evaluation \citep{DBLP:journals/corr/fed, DBLP:journals/corr/dynaeval, li-etal-2021-conversations, Yeh2021comprehensive}, as shown in Table \ref{tab:results}.
From these results, the following four conclusions can be drawn.

\noindent\textbf{\method{} Reaches Comparable Performance.}
Across three datasets, our \method{} achieves the best or comparable correlations with human evaluation.
On Controllable Dialogue dataset, all baseline metrics fail to reach meaningful correlation , while \method{} becomes the top performer.
On the other two datasets, the results of \method{} are comparable with most baselines, though the gap to the best method is obvious.
We perform an ablation study on Controllable Dialogues to further demonstrate the effectiveness of segment acts and our consensus-based framework.
Detailed description and results are documented in the Appendix~\ref{sec:appendix-ablation}.
We also list one success case and one failure case in the Appendix~\ref{sec:appendix-case-analysis} to enable a closer observation of our approach. 

\noindent\textbf{Automatic Evaluation Metrics Lack Transferability.}
We can observe that the best method on one dataset becomes mediocre on the other datasets, including our \method{}.
\method{} outperforms all other methods on Controllable Dialogue dataset, but can only get to the second tier on the other two datasets.
DynaEval, the best method on FED dataset, loses its advantage when tested on other datasets. 
The same story also happens for Flow score, a state-of-the-art metric in the DSTC9 dataset.
This observation is consistent with study from previous work \cite{Yeh2021comprehensive}.

One reason for the brittleness of these methods is that their calculations rely on large models.
The data used to train these large models plays an decisive role, as we can see from the performance difference between DynaEval\_emp and DynaEval\_daily.
In addition, \method{} depends on the segment act labels and these labels on FED dataset and DSTC9 dataset are annotated by a trained classifier.
Even though the classifier has relatively high accuracy (90\%), it still injects some errors to the segment act flow, which hinders the application of \method{} on new datasets.
These observations indicate that how to construct a robust dialogue evaluation metric remains a problem for the community.

\noindent\textbf{\method{} Can Provide Complementary Information to Other Methods.}
Similar to \citet{Yeh2021comprehensive}, we test different combinations of metrics by directly averaging one metric with the previous best metrics on the three datasets, which are BERTScore on Controllable Dialogue dataset, DynaEval\_emp on FED dataset, and Flow score on DSTC9 dataset.
The last 4 rows of Table~\ref{tab:results} show that \method{} can consistently push the current correlation ceiling to a new level the most, while many other combinations improve little or even hurt performance.
These results imply that segment act is an important missing aspect in dialogue evaluation that worth even further exploration in the future.


\noindent\textbf{Our Consensus-Based Framework Shows Potential.}
In our consensus-based framework, the retrieval step of \method{} could find pseudo-references for other reference-based metrics like BLEU \citep{papineni-etal-2002-bleu} and BERTScore \citep{DBLP:conf/iclr/ZhangKWWA20} and make them reference-free.

Here we experiment with BERTScore, as it is the best performing reference-based metric on Controllable Dialogue.
The reference-free form of BERTScore, called \emph{Consensus BERTScore}, is similar to our \method{}, except that we do not employ segment act features in the retrieval step and we exclude the segment act score, i.e., Eq.~\ref{eqn:fi}, in the assessment step.
As shown in the third row of Table~\ref{tab:results}, Consensus BERTScore slightly outperforms BERTScore in all three correlations (0.284 vs. 0.282, 0.240 vs. 0.214, 0.183 vs. 0.162).

This promising result shows the potential of our consensus-based framework. It leads a new way to rethink the usability of reference-based metrics in dialogue evaluation.

\subsection{What Does Segment Act Bring to Dialogue Evaluation?}
Compared with semantic-meaning-focused metrics, what does segment act bring to dialogue evaluation?
We hypothesize the explicit involvement of segment acts can bring useful information, complementary to semantic-meaning-focused metrics.

\begin{figure*}
    \centering
    \includegraphics[scale=0.39]{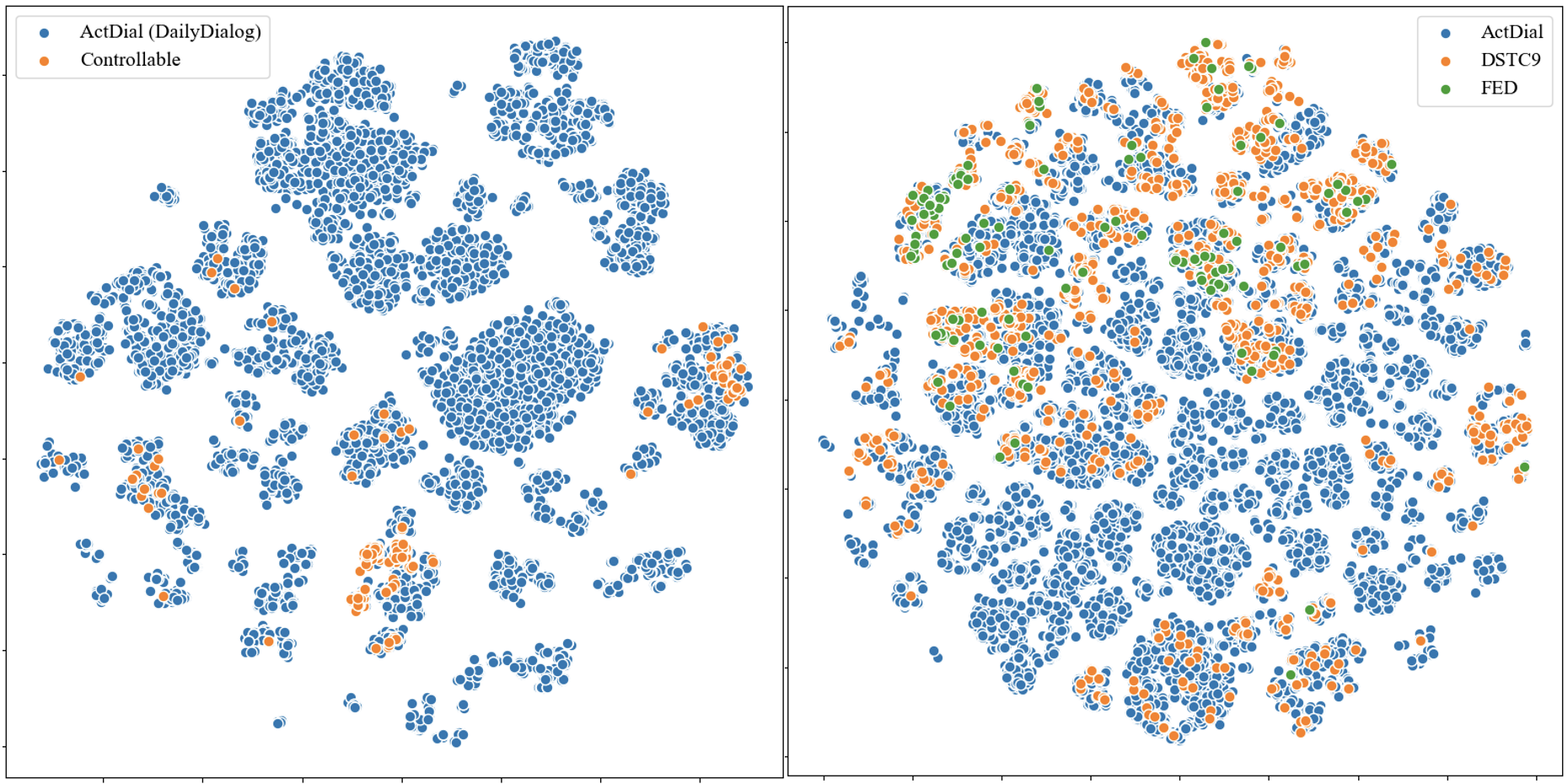}
    \caption{Segment act feature space of Controllable Dialogue, FED, DSTC9 dataset and the retrieval set \dataset{}. We have a separate plot for Controllable Dialogue because the \dataset{} we used are different (See Section~\ref{sub-sec:methods}).}
    \label{fig:dist}
\end{figure*}

\begin{figure}
\begin{center}
    \includegraphics[scale=0.4]{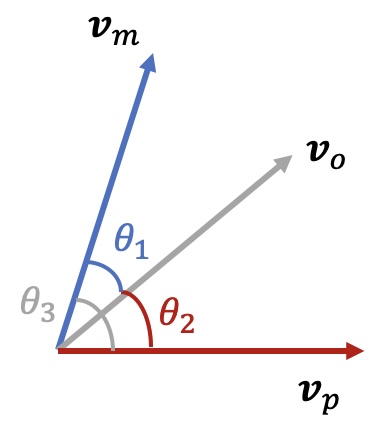}
\end{center}
\caption{
\label{fig:rank-vec}
The relationships, in our hypothesis, between human evaluation $\vec{v}_o$, semantic-meaning-focused evaluation $\vec{v}_m$, and segment-act-based evaluation $\vec{v}_p$.
}
\end{figure}

We illustrate our hypothesis in Figure~\ref{fig:rank-vec}.
If segment act is useful, the segment-act-based evaluation $\vec{v}_p$ should be positively correlated to human evaluation $\vec{v}_o$, i.e., $\vec{v}_p$ has roughly the same direction as $\vec{v}_o$ but with a small angle $\theta_2$.
If segment act is complementary to semantic-meaning-focused metrics, the segment-act-based evaluation $\vec{v}_p$ should be almost orthogonal to the semantic-meaning-focused evaluation $\vec{v}_m$, i.e., $\vec{v}_m$ falls into the other side of $\vec{v}_o$ so that $\vec{v}_m$ is also positively correlated to $\vec{v}_o$ with a small angle $\theta_1$ but almost orthogonal to $\vec{v}_p$ with a large angle $\theta_3=\theta_1+\theta_2$.
These angles $\theta_1,\theta_2$ and $\theta_3$ could be characterized by the correlation of two evaluation results.
A higher correlation implies a smaller angle.

\begin{table}[t!]
  \centering
  \setlength{\tabcolsep}{2.5pt}
  \small
  \begin{tabular}{llrrr}
    \toprule
    \makecell[c]{\textbf{Metric 1}} & \makecell[c]{\textbf{Metric 2}} & \makecell[c]{\textbf{Pearson}} & \makecell[c]{\textbf{Spearman}} & \makecell[c]{\textbf{Kendall}}\\
    \midrule
    \method{}\_seg & Human & 0.191 & 0.151 & 0.113 \\
    BLEU & Human & 0.132 & 0.136 & 0.104 \\
    BERTScore & Human & 0.282 & 0.214 & 0.162 \\
    \midrule
    \method{}\_seg & BERTScore & 0.014$^*$ & -0.030$^*$ & -0.022$^*$ \\
    \method{}\_seg & BLEU & 0.067$^*$ & 0.042$^*$ & 0.026$^*$ \\
    BLEU & BERTScore & 0.576 & 0.637 & 0.460 \\
    \bottomrule
  \end{tabular}
  \caption{\label{table:complementary}
  Inter-correlations between \method{}\_seg, BERTScore, BLEU, and human evaluation. \method{}\_seg is a version of \method{} using segment act flow only for assessment. All values are statistically significant to $p$ < 0.05, unless that are marked by $^*$.
  }
\end{table}

We conduct experiments on the test set of Controllable Dialogue dataset to validate our hypothesis.
Two of the popular semantic-meaning-focused metrics are BERTScore \cite{DBLP:conf/iclr/ZhangKWWA20} and BLEU \cite{papineni-etal-2002-bleu}.
We modify the retrieval and assessment parts of our \method{}, so that only segment act information is utilized.
We denote this variant as \emph{\method{}\_seg}.

As we could observe from the first three rows of Table \ref{table:complementary} that the \method{}\_seg, BLEU and BERTScore all exhibit strong correlation to human evaluation.
Unsurprisingly, BLEU and BERTScore are highly correlated (the last row of Table~\ref{table:complementary}), since both of them focus on the semantic meaning of dialogues.
In line with our hypothesis, the BLEU-\method{}\_seg correlation and BERTScore-\method{}\_seg correlation is far smaller (rows 4-5 of Table~\ref{table:complementary}), which indirectly shows that segment act can evaluate dialogues from a complementary perspective.
These findings resonate with the theory from \citet{bender-koller-2020-climbing}, where the \emph{meaning} and the \emph{communicative intent}, i.e., segment act here, are considered to be two decoupled and complementary dimensions.



\subsection{Why Does Consensus Work?}
We investigate why consensus-based framework can perform well in dialogue evaluation by visualizing the segment act feature space, an essential aspect in the retrieval process of \method{}.
We compare the segment act feature distribution between the three test sets and their corresponding retrieval sets, projecting these features to 2-dimensional space by t-SNE \citep{JMLR:v9:tsne} as shown in Figure~\ref{fig:dist}.
We did not tune any hyperparameter to obtain these results, in consideration of the sensitivity of t-SNE plots.

The core idea of consensus lies on using the nearest neighbors as references to measure a newcomer.
Only if the suitable nearest neighbors consistently exist, will the consensus of them have meaningful indication to evaluate a new subject.
We can observe from Figure~\ref{fig:dist} that, even though dialogues in three test sets are diverse, every datapoint from the test sets is surrounded by datapoints from the retrieval sets.
We can always reliably find good references for a new dialogue, which explains why using consensus in dialog evaluation is promising.
Moreover, this desirable coverage is achieved by an attainable amount of datapoints.
It only needs 10,494 and 31,993 dialogues as retrieval sets in our experiments to get good results.
The power of the consensus may go stronger and more reliable if the size of retrieval set grows, which could be a favorable property in many of industrial applications.



\section{Conclusion}
\label{sec:conclusion}
In this work, we propose a consensus-based reference-free framework for open-domain dialog evaluation with segment act flows. From extensive experiments against the state-of-the-art baselines, our method can reach the best or comparable correlation with human evaluation.
Our segment-act-based methods complement well to previous semantic-meaning-focused methods, pushing the ceiling of correlations. Moreover, the promise of our consensus-based framework encourages us to step further in the direction of dialog evaluation.

\section*{Limitations}

Our segment act dataset, \dataset{}, is constructed based on two widely-adopted open-domain dialogue datasets, ConvAI2 \citep{dinan2020second} and DailyDialog \citep{li2017dailydialog}.
Despite of various benefits, \dataset{} also inherits some of the limitations from ConvAI2 and DailyDialog.
The scale of the dataset could be larger.
The nature of ConvAI2 dialogues, learning personas of each other, pushes the segment act distribution towards \texttt{question} and \texttt{inform} slightly.
These limitations do not interfere too much with our methods and our extensive experiments still show significant results.
We will potentially improve our dataset in the future.

This work also brings the consensus-based framework into open-domain dialogue evaluation.
We show the effectiveness of this framework when incorporating segment act flow and content information.
Yet, the full potential of the consensus-based framework still needs more exploration.
We will leave this as future work.

\section*{Ethics Statement}
\label{sec:ethical}
A big part of this work contains (1) the data annotation on two existing benchmark datasets of conversation modeling: the ConvAI2 dataset and the DailyDialog dataset and (2) human evaluation on the overall quality of generated conversations.
As our \dataset{} is built upon the existing datasets, we follow the original copyright statements of these two datasets and will further release our segment act annotations to the research community.
During annotation, we only collected the segment act annotations, and no demographic or annotator's identity information was collected.
In addition, we provide a detail description of human evaluation design in Appendix \ref{sec:appendix-h2beval}.

\section*{Acknowledgements}
This work was supported by Centre for Perceptual and Interactive Intelligence Limited, UGC under Research Matching Grant Scheme, the Research Grants Council of the Hong Kong Special Administrative Region, China (No. CUHK 14210920 of the General Research Fund), and Tencent Rhino-Bird Research Award.

\bibliography{anthology,custom}
\bibliographystyle{acl_natbib}

\appendix

\section{Details of Our \dataset{} Dataset}
\label{sec:appendix-annotation}
\begin{table*}[t!]
  \centering
  \small
  \begin{tabular}{m{0.1\textwidth}p{0.15\textwidth}p{0.29\textwidth}p{0.2\textwidth}p{0.11\textwidth}}
    \toprule
    \makecell[c]{\textbf{General}\\\textbf{Dimension}} & \makecell[c]{\textbf{Segment Act}} & \makecell[c]{\textbf{Definition}} & \makecell[c]{\textbf{Examples}} & \makecell[c]{\textbf{Distribution}} \\
    \midrule 
    \multicolumn{1}{l}{\multirow{4}{*}[-1.5cm]{\makecell[l]{General\\Task}}} &
    \texttt{inform} & The sender makes the addressee know some information which he assumes to be correct. & ``The train is leaving.'', \newline ``The meeting starts in 5 minutes.'' & \makecell[r]{65.702\%} \\
    & \texttt{question} & The sender asks the addressee to provide some information which he assumes the addressee knows. & ``What time is it?'', \newline ``Where is the nearest bank?'' & \makecell[r]{16.529\%} \\
    & \texttt{directive} & The sender asks the addressee to perform an action. & ``Please don’t do this ever again.'' & \makecell[r]{2.880\%} \\
    & \texttt{commissive} & The sender considers to perform an action which he believes would be in addressee’s interest,  or he has been requested/suggested to perform by the addressee. & ``I will not do that any more.'', \newline ``May I offer you an upgrade?'' & \makecell[r]{0.517\%} \\
    \midrule
    \multirow{4}{*}[-0.5cm]{\makecell[l]{Social\\Obligation\\Management}} & \texttt{greeting} & The speakers inform the presence of each other. & ``how are you?'', \newline ``I'm fine.'' & \makecell[r]{6.023\%} \\
    & \texttt{goodbye} & The speakers inform the end of the dialog. & ``Bye'', ``See you.'' & \makecell[r]{0.172\%}  \\
    & \texttt{apology} & The speakers express or mitigate the feelings of regret. & ``Sorry.'', ``No problem.'' & \makecell[r]{0.542\%} \\
    & \texttt{thanking} & The speakers express or mitigate the feelings of gratitude. & ``Thanks.'', \newline ``You are welcome.'' & \makecell[r]{1.049\%} \\
    \midrule
    \multirow{3}{*}[-0.5cm]{\makecell[l]{Simple\\Feedback}} & \texttt{backchannel-}\newline\texttt{success} & The speakers succeed in processing the previous dialog. & ``Okay'', ``Uh-huh'' & \makecell[r]{6.543\%} \\
    & \texttt{backchannel-}\newline\texttt{failure} & The speakers fail in processing the previous dialog. & ``Sorry?'', ``Excuse me?'' & \makecell[r]{0.030\%} \\
    & \texttt{check-}\newline\texttt{understanding} & The sender wants to check whether the addressee succeed in processing the previous dialog. & ``Do you get what I just said?'' & \makecell[r]{0.013\%} \\
    \bottomrule
  \end{tabular}
  \caption{\label{tab:tag-definition}
  Our ISO-format open-domain segment act tagset: the definition, examples, and distribution}
\end{table*}

\subsection{Segment Acts definitions}
\label{sec:appendix-intent-definition}
For the formal definitions and examples of segment act, please refer to Table \ref{tab:tag-definition}.
The eleven segment act labels cover three major communication activities: (i) general task, which includes information-transfer activities and action-discussion activities; (ii) social obligation management, which includes typical social conventions in communication; and (iii) simple feedback, which includes simple non-informative feedback about the processing of previous utterances. 

We segmented all the dialogue utterances using the NLTK sentence punctuation tokenizer \citep{kiss-strunk-2006-unsupervised} that mainly consists of a set of rule-based regular expressions on punctuation.

\subsection{Crowdsourcing Segment Act Annotation}
\label{sec:appendix-dataset-crowdsource}
We crowdsourced segment act annotation from annotators whose native language is Mandarin Chinese (zh-cmn), but more importantly, they are proficient in English (en-US). 
More than 50 annotators participated after rigorous training to ensure data quality. 
Each segment is annotated by three different annotators.
If the initial three annotations are all different, further round(s) of annotation on this segment would be conducted until it got a majority vote (at least two annotations are the same).

Besides Fleiss' kappa \citep{fleiss1971measuring} mentioned in Section \ref{sec:dataset}, we here report Fleiss' kappa in a new setting and the overall sample accuracy to show the quality of our annotations. 

Since the segment act distribution is unbalanced, we calculated another Fleiss’ kappa excluding all the annotations with the most dominant segment act, i.e., \texttt{inform}, to eliminate potential bias.
In this setting, the new kappa is 0.768 for DailyDialog and 0.775 for ConvAI2, staying roughly the same as the overall ones.
These results prove the robustness of our annotations.

Although it is impossible to check the correctness of every single annotation, we do perform sampling inspection when collecting the annotations everyday. 
In total, We sampled 8,000 segments randomly and annotated these segments by ourselves. 
Since we have a deeper understanding than our annotators and our annotations are examined multiple times by ourselves, our annotations on these 8,000 segments can be considered as ground truth.
The majority votes of crowdsourced annotations are later compared with the ground truth labels to obtain sample accuracy. The sample accuracy in DailyDialog annotation is 0.90 and that in ConvAI2 is 0.93.
The small gap of the accuracy is due to the difference in dialogue complexity.

\subsection{Dataset Statistics and Distributions}
\label{sec:appendix-dataset-statistics}
For the ConvAI2 dataset, we collected 481,937 segment acts on the training set, and 29,232 segment acts on the validation set. 
Since the testing set is not publicly available, we did not annotate it.

For the DailyDialog dataset, we gathered 178,604 segment acts on the training set, 16,500 segment acts on the validation set, and 16,028 segment acts on the testing set.

Note that even though ConvAI2 and DailyDialog split their data for training, validation, and testing purpose, it is not always necessary to mechanically follow the splits.
Our annotations on ConvAI2 and DailyDialog can be used as a unity, \textit{\dataset{}}, depending on the research problems.

Table \ref{tab:tag-definition} shows the distribution of all segment acts on our dataset.
The segment act distribution is unbalanced.
Specifically, the distribution is highly skewed to \texttt{inform} and \texttt{question}, which is not surprising because ConvAI2 and DailyDialog are chitchat datasets and the majority of communication activities is exchanging information. 
In addition, few written dialogues between two strangers, the setting of ConvAI2 and DailyDialog, involve apology or encounter communicative difficulties, which results in the rare occurrences of \texttt{apology}, \texttt{backchennel-failure}, and \texttt{check-understanding} segment acts.
However, it is still essential to include these segment acts as they take place more commonly in spoken dialogues in the real world.

\section{Ablation Study on Controllable Dialogues}
\label{sec:appendix-ablation}
We perform ablation study on Controllable Dialogues and obtained positive results. 
This experiment is designed to reveal the effectiveness of segment act, so content-related information and features are excluded from the whole process.
Specifically, we remove the content feature and only used the segment act flow feature during the retrieval (Section \ref{sec:retrieval}). 
We later assessed each dialogue on this shrunk retrieval set. 
The Pearson, Spearman’s, and Kendall correlations in this setting are 0.298, 0.252, and 0.189 respectively. 
These results decrease slightly from our full version of FlowEval (0.301, 0.256, and 0.193) but remain higher than the previous SOTA (0.282, 0.214, and 0.162). 
This ablation study strengthens our claim on the effectiveness of segment acts and our consensus-based framework.

\section{Human Evaluation for Controllable Dialogue}
\label{sec:appendix-h2beval}
\begin{table*}[ht]
  \centering
  \small
  \begin{tabular}{lccccc}
    \toprule
    &\multicolumn{4}{c}{\textbf{Controllable Dialogue Dataset}} \\
    \cmidrule(lr){2-6}
    & \textsc{Relevancy} & \textsc{No Contradiction} & \textsc{No Repetition} & \textsc{Consistency} & \textsc{Overall}\\
    \midrule
    \textbf{Seq2Seq} & 0.7802 & 0.8791 & 0.3846 & 0.9010 & 3.4505 \\
    \textbf{Seq2Seq + Repetition} & \textbf{0.8437} & \textbf{0.9062} & 0.7812 & 0.8541 & 3.6250 \\
    \textbf{Seq2Seq + Specificity} & 0.8351 & 0.8681 & \textbf{0.8351} & \textbf{0.8681} & \textbf{3.8791} \\
    \bottomrule
  \end{tabular}
  \caption{\label{tab:human-bot-result}
  Controllable Dialogue \citep{see2019makes} evaluation results by AMT crowd-workers.
  }
\end{table*}
We collected human judgements from Amazon Mechanical Turk (AMT).
The crowd-workers are provided with the full multi-turn conversation for evaluation.
We ask crowd-workers to evaluate the \textit{relevancy}, \textit{avoiding contradiction}, \textit{avoiding repetition}, \textit{persona consistency} and \textit{overall quality} of the conversation.
The reason for designing the human evaluation on different aspects is that we assume a good conversation between human and {a dialogue system} should satisfy the following properties: 
(1) generating relevant and non-repetitive responses (\textit{relevancy} and \textit{avoiding repetition}), 
(2) memorizing the dialogue history and generating non-contradictory information (\textit{avoiding contradiction}), 
(3) maintaining a consistent persona/topic (\textit{persona/topic consistency}),
(4) formulating a natural conversation (\textit{overall quality}).

The first four aspects are formulated as binary-choice questions, and the overall quality is formulated as Likert question on a 1-5 scale, where higher is better.
{During} evaluation, we did not distinguish whether {an} utterance is generated by human or by dialogue model, because we want {the evaluation is about} the full conversation, rather than just utterances generated by the dialogue model.

To ensure better data quality, Turkers are selected by their job success rate and geographic location (only admits turkers from English speaking countries). Before starting our evaluation job, turkers must read through our detailed guideline.
For each dialogue, a turker is asked to evaluate the dialogue from the following perspectives:
\begin{enumerate}
    \item \textbf{Irrelevant response (binary)}: Whether or not the speaker generates a response which seems to come out of nowhere according to the conversation history. Binary score.
    \item \textbf{Contradictory information (binary)}: Whether or not the speaker generates a response which contradicts to the common sense or to what himself just said in the previous conversation. Binary score.
    \item \textbf{Repetitive response (binary)}: Whether or not the speaker generates a response which has the same meaning as his previous utterance(s). Binary score.
    \item \textbf{Inconsistent with persona (binary)}: Whether or not the speaker generates a response which is not consistent with his persona profile. \textbf{Only used if the dialogues-to-evaluate follow ConvAI2 setting and are generated with personas.} Binary score.
    \item \textbf{Topic shifts (binary)}: Whether or not the speaker generates a response which belongs to a completely different topic compared with the previous conversation history. \textbf{Only used if the dialogues-to-evaluate follow Daily Dialogue setting and are not generated with personas.} Binary score.
    \item \textbf{Overall score (1-5)}: An overall impression of the dialogue quality, not necessary to have any relationship with the aspects above. Score is between 1 to 5 inclusive, all integer. The higher the better.
\end{enumerate}

The evaluation results are examined by ourselves. Incorrect annotation would be rejected and re-evaluated by another turker.
The final evaluation results are shown as Table \ref{tab:human-bot-result}.


\section{Implementation of ActBERT}
\label{sec:appendix-actbert}
ActBERT follows the architecture of RoBERTa \citep{DBLP:journals/corr/abs-1907-11692}.
The vocabulary size is relatively small as it only contains 11 segment acts and other special tokens.
It has 4 hidden layers, 4 attention heads, and a hidden dimension size of 256.
Speaker information is included using different input token type.
Similar to the masked language model task, we use a masked segment act task during the training.

\section{Implementation of BLEU and BERTScore}
\label{sec:appendix-bleu&bertscore}
Controllable Dialogue \citep{see2019makes} are trained on ConvAI2 dataset whose setting is two participants talking about their own personas.
This unique characteristics makes it feasible to find references for BLEU, BERTScore, or other reference-based metrics.

We take dialogues, from the testing set of ConvAI2, that have the most overlapping personas as the references for a dialogue.
Although not as convincing as references in machine translation task, references obtained in this way prove to be helpful to dialogue evaluation.
Both BLEU and BERTScore reaches relatively high correlations on Controllable Dialogue.

The smooth function of the BLEU score is NIST geometric sequence smoothing \citep{bird-loper-2004-nltk}. BERTScore are calculated by using the package from its authors \citep{DBLP:conf/iclr/ZhangKWWA20}.

\section{Cases Analysis}
\label{sec:appendix-case-analysis}
\begin{table}[t!]
  \centering
  \setlength{\tabcolsep}{5pt}
  \small
  \begin{tabular}{p{6.5cm}}
    \toprule
    \textbf{Human}: My motorbike is giving me trouble. What a bad start to the day! \\
    \midrule
    \textbf{Chatbot}: Sorry to hear that. What do you do for a living? \\
    \midrule
    \textbf{Human}: I'm a waitress. What about you? \\
    \midrule
    \textbf{Chatbot}: I am old, old, old, old, old, old, old, old, old, old, old, old. \\
    \midrule
    \textbf{Human}: Ha ha! Ok. I'm young with big brown eyes always looking for fun! \\
    \midrule
    \textbf{Chatbot}: I am old, old, old, old, old, old, old, old, old, old, old, old, old, old. \\
    \midrule
    \textbf{Human}: I get it!! So do you like mtorbikes? \\
    \midrule
    \textbf{Chatbot}: I am old, old, old, old, old, old, old, old, old, old, old, old, old, old, old. \\
    \midrule
    \textbf{Human}: Again. Can't you say anything else? \\
    \midrule
    \textbf{Chatbot}: Old, old, old, old, old, old, old, old, old, old, old, old, old, old, old, old, old. \\
    \midrule
    \textbf{Human}: You already said that. \\
    \midrule
    \textbf{Chatbot}: Old, old, old, old, old, old, old, old, old, old, old, old, old, old, old, old, old. \\
    \bottomrule
  \end{tabular}
  \caption{\label{table:success-case}
  A conversation from Controllable Dialogue with human rating as 1 (lowest). \method{} can successfully give it the lowest score, while DynaEval, Flow score fail to distinguish the low quality.
  }
\end{table}
\begin{table}
  \centering
  \setlength{\tabcolsep}{5pt}
  \small
  \begin{tabular}{p{6.5cm}}
    \toprule
    \textbf{Human}: Hello how are you today \\
    \midrule
    \textbf{Chatbot}: Doing well, how about you? \\
    \midrule
    \textbf{Human}: I'm doing well just getting ready to go to work \\
    \midrule
    \textbf{Chatbot}: Nice, where do you work? \\
    \midrule
    \textbf{Human}: I am a librarian what about you \\
    \midrule
    \textbf{Chatbot}: I work as an administrator at the office of a local construction company. \\
    \midrule
    \textbf{Human}: That sounds like a lot of fun \\
    \midrule
    \textbf{Chatbot}: Oh, believe me, it's not. So, do you have any hobbies? \\
    \midrule
    \textbf{Human}: I love being in the library \\
    \midrule
    \textbf{Chatbot}: It's good to really enjoy your work. \\
    \midrule
    \textbf{Human}: Thank you very much \\
    \midrule
    \textbf{Chatbot}: So, have you ever travelled outside the country? \\
    \bottomrule
  \end{tabular}
  \caption{\label{table:failure-case}
  A conversation from Controllable Dialogue with human rating as 4 (high quality). \method{} and Flow score fail to give it a high score, while DynaEval considers it one of the best.
  }
\end{table}
In this section, we exhibit two real cases from Controllable Dialogue to show the strengths and weaknesses of our \method{}.

\paragraph{A Successful Case}
Table~\ref{table:success-case} shows a poor conversation with human rating 1 point, the lowest point possible. Our \method{} can successfully rank this dialogue as the worst among 278 dialogues, while DynaEval and Flow score consider it not bad but mediocre.
To be exact, DynaEval ranks it at the top 170th and Flow score places it at the 114th.

Repetition is the main flaw of this conversation.
From the segment act feature perspective, the abnormal long \texttt{inform} flows caused by a series of ``old'' provide \method{} enough evidence to penalize this behavior.
From content feature aspect, the repeated words do not contribute well to our word overlapping assessment in \method{} either.
Other methods do not have specific mechanism that help them distinguish this repetition, which makes them vulnerable in similar cases.

\paragraph{A Failure Case}
Here we show an example where \method{} fails to deliver the correct evaluation.
The dialogue in Table~\ref{table:failure-case} is of high quality: annotators give it 4 points, the highest points among 278 dialogues.
However, \method{} ranks it at the 253rd. and Flow score ranks it at the 178th.
DynaEval gets it right, reviewing it as the 3rd best dialogue.

The segment act flow of this dialogue is relatively natural.
The contents appear not so frequently in the \dataset{} dataset. 
These characteristics make it hard for \method{} to output a correct ranking.

Just like our competing baselines, more analysis and case study are needed to determine a more concrete pattern.



\section{Computational Cost}

All of our experiments are run on a single NVIDIA V100 GPU.
Note that our method does not require excessive computational power and GPU with lower computational ability can reproduce our results in reasonable amount of time.

\end{document}